\documentclass[sigconf]{acmart}

\usepackage{amsmath,bm}
\usepackage[linesnumbered,ruled,vlined]{algorithm2e}
\usepackage{cuted}
\usepackage{booktabs}
\usepackage{multirow}
\usepackage{tabularx}
\usepackage{makecell}
\usepackage{graphicx}
\usepackage{xcolor}
\usepackage{enumitem}
\usepackage{tikz}
\usepackage{pgfplots}
\usepackage{subcaption}
\usepackage{placeins}

\pgfplotsset{compat=1.18}
\usetikzlibrary{arrows.meta,backgrounds,calc,fit,matrix,positioning,shapes.geometric}

\definecolor{mobiblue}{RGB}{42,105,176}
\definecolor{mobigreen}{RGB}{50,145,96}
\definecolor{mobiorange}{RGB}{224,132,55}
\definecolor{mobired}{RGB}{196,68,68}
\definecolor{mobipurple}{RGB}{111,78,154}
\definecolor{mobilightblue}{RGB}{225,238,250}
\definecolor{mobilightgreen}{RGB}{226,244,233}
\definecolor{mobilightorange}{RGB}{252,237,219}
\definecolor{mobilightgray}{RGB}{242,244,247}

\newcommand{\name}{\textsc{MobiWave}}

\newcommand{\R}{\mathbb{R}}

\newcommand{\E}{\mathbb{E}}

\setcopyright{none}
\copyrightyear{2027}
\acmYear{2027}
\acmConference[KDD '27]{The 33rd ACM SIGKDD Conference on Knowledge Discovery and Data Mining}{August 2027}{San Jose, CA, USA}
\acmBooktitle{Proceedings of the 33rd ACM SIGKDD Conference on Knowledge Discovery and Data Mining (KDD '27)}
\acmDOI{}
\acmISBN{}

\AtBeginDocument{}

\begin{document}

\title{MobiWave: Dispatch-Oriented Graph Wavelets and Drift-Guided Selective Optimization for Autonomous Fleet Rebalancing}

\author{Xiao Han, Pinbo Wang, Yuanshao Zhu, Guojiang Shen, Xiangjie Kong}
\renewcommand{\shortauthors}{X. Han, P.B. Wang, Y.S. Zhu, G.J. Shen, and X.J. Kong}

\begin{abstract}
Autonomous fleets enable mobility platforms to coordinate idle vehicles directly, making fleet-wide rebalancing possible.
However, two obstacles limit reliable deployment: overlapping regional and local traffic patterns can hide roads that remain useful for dispatch, and mobility drift can make a trained policy unreliable.
Existing spatial aggregation mixes these patterns, while updating all parameters from limited recent data is costly and can damage stable knowledge.
We propose \name, a framework that connects a dispatch-oriented multi-scale graph wavelet module with Drift-Guided Layer-Selective Optimization (DGLS).
The first module addresses the representation challenge by separating graph-frequency patterns and weighting each scale according to its value for demand prediction and feasible rebalancing.
DGLS addresses the adaptation challenge by measuring Dispatch-weighted Spectral Drift, selecting affected layers within a resource budget, and separating short shocks from persistent changes through a drift-aware fast--slow update.
Candidate validation further rejects updates that fail to improve held-out dispatch reward without worsening monitored service or safety constraints.
Experiments on both real-world datasets and simluated environments demonstrate the effectiveness of \name\ in comparing with state-of-the-art methods.
The source code and datasets are available at \url{https://anonymous.4open.science/r/MobiWave-40F8/}.

\end{abstract}



\maketitle

\section{Introduction}

\begin{figure}[htb!]
\centering
\begin{subfigure}[t]{0.42\linewidth}
\centering
\includegraphics[width=\linewidth]{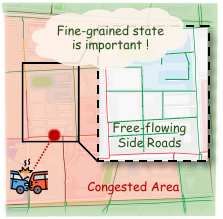}
\caption{Fine-Grained Road States}
\label{fig:1a}
\end{subfigure}
\begin{subfigure}[t]{0.505\linewidth}
\centering
\includegraphics[width=\linewidth]{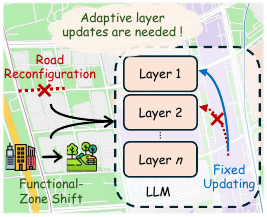}
\caption{Adaptive Layer Updates}
\label{fig:1b}
\end{subfigure}
\caption{Motivation for \textsc{\name}: (a) Broad congestion can hide free-flowing side roads that are useful for rebalancing. (b) Road reconfiguration and functional-zone shifts can affect different model layers, while fixed updates may miss them.}
\Description{The left panel contrasts a congested area with nearby free-flowing side roads. The right panel shows road and functional-zone changes pointing to different model layers, while a fixed update misses the affected layer.}
\label{fig:1}
\end{figure}

Urban mobility platforms commonly assign incoming requests to nearby drivers~\cite{lin2018fleet,chen2024irebalance,han2025garlic}.
Because drivers may favor familiar areas or particular trips, the resulting dispatch is not fully controlled by the platform \cite{chen2024irebalance,li2025learning}.
The platform must therefore model network-wide supply--demand conditions as well as drivers' responses to dispatch instructions.
Autonomous vehicles change this setting because they can carry out feasible platform decisions directly, making driver behavior no longer the main source of uncertainty~\cite{pavone2012robotic,zhang2016control}.
Dispatch can then shift from driver-constrained order matching to fleet-wide rebalancing that maximizes long-term system value.
This shift makes two abilities critical: extracting road-network information that supports rebalancing \cite{li2025integrated,wang2025merci} and maintaining reliable dispatch decisions \cite{yang2025large} as urban mobility conditions evolve.

Both abilities are difficult because urban mobility changes at several scales.
Regional supply and demand follow hourly, daily, and weekly cycles, and their changes can spread across connected regions \cite{han2025garlic}.
Accidents, events, and temporary construction can instead produce sharp local changes \cite{zhang2025matador}.
These patterns overlap in space and time, making it difficult to identify their distinct effects on fleet decisions.
For example, a city-wide rush hour and a short road closure can both create nearby vehicle shortages, although they call for different rebalancing actions.
Over longer periods, new roads and urban rezoning reshape connectivity, while changes in residents' travel habits shift regional demand \cite{la2026habit}.
These changes weaken models trained on historical data.
Retraining on the complete history is costly, whereas updating every parameter from a short recent window can be unsafe.
We therefore ask how an autonomous fleet can extract dispatch-critical road-network information and adapt its policy from limited recent data.

The first part of this question concerns the road-network representation provided to the dispatcher.
Existing end-to-end spatiotemporal models commonly mix road-network signals through repeated aggregation over neighboring regions~\cite{li2018dcrnn,yu2018stgcn,bai2019stg2seq}.
Figure~\ref{fig:1}(a) illustrates how this aggregation can blur local conditions.
Although congestion appears to spread around the accident site, many nearby side streets remain free-flowing.
Treating the entire area as congested would hide viable routes and lead to poor vehicle rebalancing.
Graph spectral methods offer a way to separate road-network signals by frequency~\cite{defferrard2016chebnet,xu2019gwnn,liu2025wavegc,han2020congestion}, and recent two-dimensional filters model joint spatial and temporal spectral relations~\cite{chen2025stsgnn}.
Yet these methods do not identify which graph-frequency components are most useful for dispatch in each region and time step.
Therefore, the first challenge is to separate overlapping road-network patterns across scales and retain the components that support vehicle rebalancing.

Accurately representing dispatch-relevant road-network conditions and maintaining a reliable dispatch model under evolving mobility conditions are equally important.
Large spatiotemporal dispatch models learn stable mobility relations from extensive historical data.
Some of these relations become outdated when roads are reconfigured, urban functions shift, or residents change where and when they travel.
A recent window better reflects current conditions, yet it is usually too small to support a safe update of every parameter.
Periodic full-model tuning is also costly in time and memory and can overfit a short abnormal window.
Distribution tests can reveal changes between recent and historical states~\cite{gretton2012kernel}, while regularization and replay can preserve earlier knowledge~\cite{kirkpatrick2017ewc,rolnick2019replay}.
Recent studies reduce adaptation cost by updating selected parameters, routing data through adaptive experts, or maintaining optimizer states at several time scales~\cite{wang2021tent,hu2022lora,maharana2025palm,zhao2026freqctta,behrouz2025nested}.
Figure~\ref{fig:1}(b) illustrates that road reconfiguration and functional-zone shifts may affect different internal representations, which a fixed update rule can miss.
A reliable update must therefore connect each observed road-network change to the model components that control the affected dispatch decisions.
Existing approaches do not jointly measure dispatch-relevant drift, choose affected layers within a resource budget, preserve stable knowledge, and reject harmful updates.
Therefore, the second challenge is to locate and update only the model components affected by current mobility drift, using limited recent data without erasing historical knowledge.

To address these challenges, we propose \name, which combines dispatch-oriented graph wavelets with drift-guided selective optimization for autonomous fleet rebalancing.
To address the first challenge, the dispatch-oriented multi-scale graph wavelet module combines causal temporal features with graph wavelets to separate city-wide trends, cross-region propagation, and local disruptions.
Dispatch-aware gating then weights these graph-frequency scales according to their value for rebalancing.
To address the second challenge, DGLS measures Dispatch-weighted Spectral Drift, selects affected layers within a resource budget, and uses drift-aware fast--slow updates and candidate validation to protect stable knowledge.
The dispatch evidence learned by the first module thus supports both current rebalancing and selective adaptation, linking the two challenges within one framework.
Our contributions are summarized as follows:
\begin{itemize}[left=0pt,topsep=2pt,itemsep=1pt,parsep=0pt,partopsep=0pt]
    \item We formulate autonomous fleet rebalancing under urban mobility drift as a joint problem of dispatch-critical state perception and continual model adaptation.
    \item We introduce a dispatch-oriented multi-scale graph wavelet module that separates road dynamics at multiple scales and learns scale weights for demand prediction and fleet control.
    \item DGLS is developed to measure Dispatch-weighted Spectral Drift, select and adapt affected layers within a fixed budget, and reject updates that fail candidate validation.
    \item Experiments on both real-world datasets and the simulation platform demonstrate the effectiveness of \name\ in reducing empty-loaded rate and improving profit.
\end{itemize}

\section{Preliminaries}
\label{sec:preliminaries}

\noindent\textbf{Road-network representation.}
We divide a city into $N$ non-overlapping dispatch zones and represent their one-step reachability at time $t$ by an undirected weighted graph $\mathcal{G}_t=(\mathcal{V},\mathcal{E}_t,\mathbf{W}_t)$, where $\mathcal{V}=\{v_1,\ldots,v_N\}$ is the zone set.  The symmetric weights combine feasible reachability, free-flow travel time, and historical bidirectional origin--destination flow.  With $[\mathbf{D}_t]_{ii}=\sum_j[\mathbf{W}_t]_{ij}$, the normalized Laplacian $\mathbf{L}_t=\mathbf{I}-\mathbf{D}_t^{-1/2}\mathbf{W}_t\mathbf{D}_t^{-1/2}$ defines graph spatial frequency; the inverse degree is set to zero for an isolated zone.  A road closure or new connection updates $\mathcal E_t$, $\mathbf{W}_t$, and $\mathbf{L}_t$.

\noindent\textbf{Traffic and fleet states.}
At dispatch time $t$, $\mathbf{X}_t\in\R^{N\times F}$ stacks the causal zone states $\mathbf{x}_{i,t}=[d_{i,t},n_{i,t},b_{i,t},s_{i,t},\boldsymbol{\tau}_{i,t},\boldsymbol{c}_{i,t},\mathbf{e}_{i,t}]$, including current demand, available vehicles, request backlog, passenger assignments, candidate-move travel times and costs, and observed external factors.  The state also contains in-transit records, which the proposed module summarizes as expected arrivals $\mathbf a_t^{\mathrm{tr}}$.
Every feature used at time $t$ is observed no later than $t$.

\noindent\textbf{Rebalancing action.}
The platform first assigns $s_{i,t}$ vehicles to passenger orders and then controls the remaining idle vehicles $\bar n_{i,t}=n_{i,t}-s_{i,t}$ through an integer flow matrix $\mathbf{U}_t=[u_{ij,t}]\in\mathbb{N}_0^{N\times N}$.  The entry $u_{ii,t}$ denotes vehicles that stay in zone $i$, while $u_{ij,t}$ for $i\ne j$ denotes vehicles sent from $i$ to an adjacent zone $j$.  Feasible actions must satisfy
\begin{equation}
 u_{ij,t}=0\ \text{if }i\ne j\text{ and }(v_i,v_j)\notin\mathcal{E}_t,\ 
 \sum_{j=1}^{N}u_{ij,t}=\bar n_{i,t}.
 \label{eq:action-feasibility}
\end{equation}

Fleet conservation additionally accounts for vehicles arriving from earlier moves and completed passenger trips; in-transit vehicles remain in the fleet state until each arrives.

\noindent\textbf{Problem definition.}
Let $r_t$ denote the one-step dispatch reward, defined as passenger fare revenue minus the costs of empty travel, passenger waiting, and canceled requests.  Given policy $\pi_{\theta}$, autonomous fleet rebalancing seeks
\begin{equation}
 \max_{\theta}\ J(\theta)=
 \E_{\pi_{\theta}}\!\left[\sum_{t=0}^{T-1}\gamma^t r_t\right]
 \quad s.t.\text{ \eqref{eq:action-feasibility} and fleet conservation},
 \label{eq:problem}
\end{equation}
where $\gamma\in(0,1]$.  We study the harder case in which the data distribution changes over time and only a small recent window and a fixed update budget are available.

\section{Methodology}
\label{sec:methodology}

\subsection{Framework Overview}
\label{sec:overview}

\begin{figure*}[t]
\centering
\includegraphics[width=0.95\textwidth]{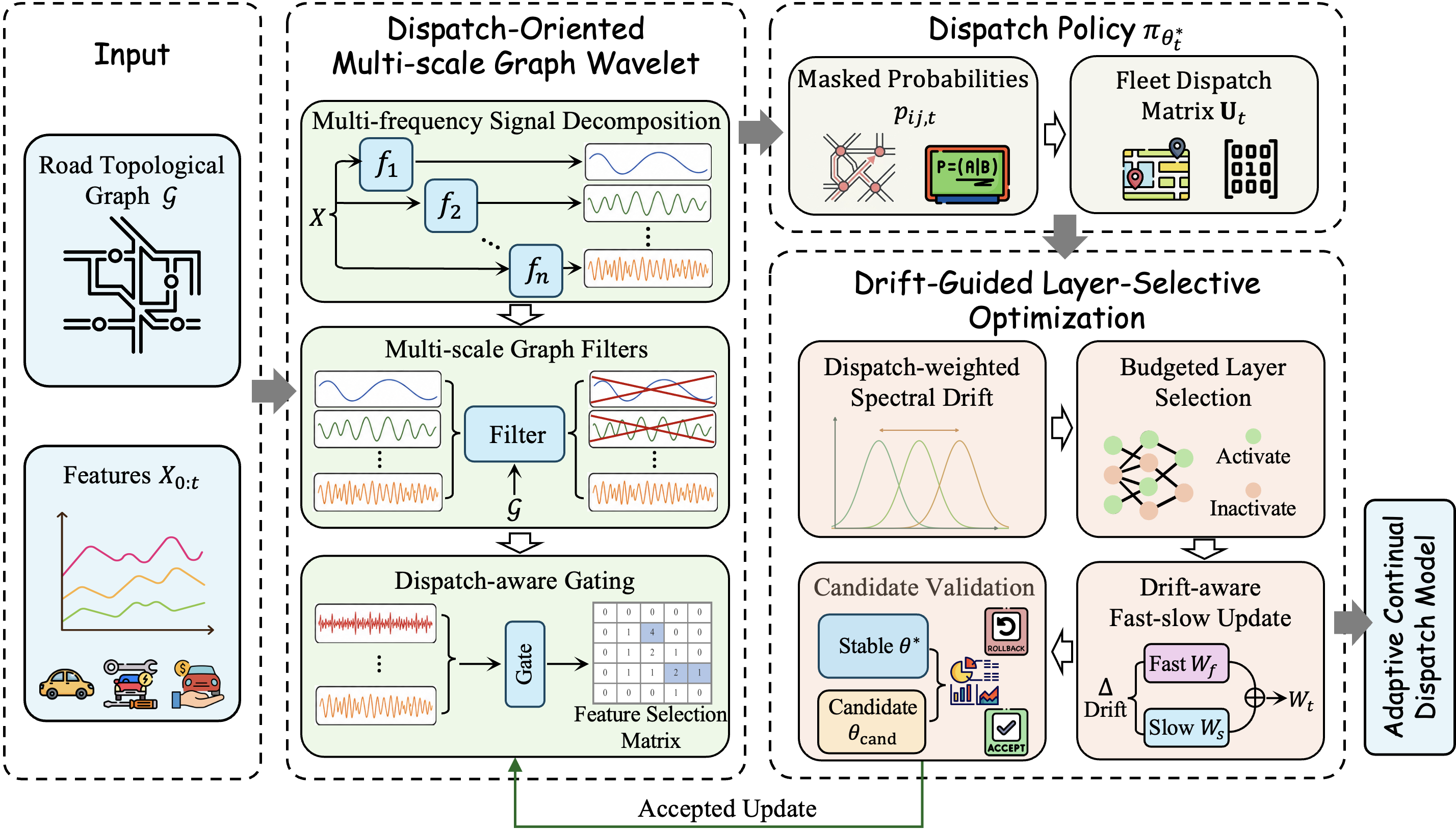}
\caption{The MobiWave framework. The dispatch-oriented multi-scale graph wavelet module supports fleet rebalancing. DGLS measures dispatch-weighted spectral drift, selectively updates affected layers under a resource budget, and applies candidate validation before deployment.}
\label{fig:overview}
\end{figure*}

\autoref{fig:overview} presents \name, which maps the road graph, causal road-network history, and current fleet distribution to feasible rebalancing decisions through two connected modules. The dispatch-oriented multi-scale graph wavelet module separates broad and local graph-frequency patterns and fuses the components that support demand prediction and rebalancing. The resulting scale features and gate weights also provide dispatch evidence for DGLS, which detects relevant mobility drift, selects and adapts affected layers under a resource budget, and validates every candidate update before deployment. The shared representation therefore supports both current dispatch and controlled adaptation.

\subsection{Dispatch-Oriented Multi-scale Graph Wavelet}
\label{sec:wavelet}

Reliable rebalancing requires a road-network representation that separates broad traffic patterns from local road changes and identifies which pattern matters at each zone and time.
Existing neighborhood aggregation can mix these signals and hide useful local conditions, whereas graph wavelets separate graph-frequency components while retaining their locations.
Combining graph wavelets with causal temporal features and dispatch-aware gating allows the model to select the components that support fleet decisions.
We therefore use a dispatch-oriented multi-scale graph wavelet module to construct and fuse scale features for fleet dispatch.

Constructing this representation from a single state is insufficient because it cannot distinguish temporary from persistent imbalance, while future-dependent summaries would invalidate online dispatch.
To expose only changes observed by time $t$, the module combines the current state, multi-horizon temporal summaries, periodic code $\mathbf p_t$, neighborhood demand--supply gap $\boldsymbol\delta_t^{\mathrm{nbr}}$, and vehicles scheduled to arrive $\mathbf a_t^{\mathrm{tr}}$:
\begin{equation}
\begin{aligned}
\mathbf Z_t
 &=\left[\mathbf X_t\middle\|
 \mathop{\|}_{h\in\mathcal H}
 [\mathbf M_t^{(h)}\|\boldsymbol\Delta_t^{(h)}]
 \middle\|\mathbf 1_N\mathbf p_t^\top
 \middle\|\boldsymbol\delta_t^{\mathrm{nbr}}
 \middle\|\mathbf a_t^{\mathrm{tr}}\right]\mathbf W_z+\mathbf 1_N\mathbf b_z^\top  ,
\end{aligned}
\label{eq:causal-input}
\end{equation}
where, for each horizon $h\in\mathcal H$, the current-window mean and its change from the immediately preceding window are
$\mathbf M_t^{(h)}
 =\frac{1}{h}\sum_{q=0}^{h-1}\mathbf X_{t-q}$,
$\boldsymbol\Delta_t^{(h)}
 =\mathbf M_t^{(h)}-\frac{1}{h}\sum_{q=h}^{2h-1}\mathbf X_{t-q}$.
The paired windows describe both the current level and its recent direction at several horizons.
Because every time index is at most $t$, Equation~\eqref{eq:causal-input} distinguishes short-lived changes from persistent imbalances without using future observations.

Although Equation~\eqref{eq:causal-input} uses only causal observations, directly aggregating $\mathbf Z_t$ over neighboring zones could still mix broad traffic patterns with fine-grained road changes.
We therefore decompose the input over multiple graph-frequency scales before dispatch-specific fusion.
Let $\mathbf L_t=\mathbf Q_t\boldsymbol\Lambda_t\mathbf Q_t^\top$ be the normalized Laplacian of the current road graph and $\mathcal B=\{1,\ldots,K\}$ be the set of spectral scales.
For each $b\in\mathcal B$, the localized graph-wavelet feature is
\begin{equation}
\mathbf H_t^b
=
\operatorname{LN}\!\left(
\operatorname{ReLU}\!\left(
g_b(\mathbf L_t)\mathbf Z_t\mathbf W_b
+
\mathbf 1_N\mathbf b_b^\top
\right)
\right),
\qquad b\in\mathcal B,
\label{eq:wavelet_feature}
\end{equation}
where $g_b(\mathbf L_t)=\mathbf Q_t g_b(\boldsymbol\Lambda_t)\mathbf Q_t^\top$ applies the kernel of scale $b$.
Given ordered heat scales $\xi_1>\cdots>\xi_{K-1}>0$, these kernels are
\begin{equation}
g_b(\lambda)
=
\begin{cases}
\exp(-\xi_1\lambda),
& b=1,\\[2pt]
\exp(-\xi_b\lambda)-\exp(-\xi_{b-1}\lambda),
& 2\le b\le K-1,\\[2pt]
1-\exp(-\xi_{K-1}\lambda),
& b=K.
\end{cases}
\end{equation}

The nonnegative kernels sum to one at every graph frequency, so the resulting features jointly cover the full spectrum.
Low-frequency scales capture smooth regional patterns, whereas high-frequency scales preserve fine-grained local changes that conventional neighborhood aggregation may obscure.

Direct eigendecomposition would be too costly whenever the road graph changes.
With $\widetilde{\mathbf L}_t=\mathbf L_t-\mathbf I$, all scales instead reuse a shared Chebyshev basis:
\begin{equation}
g_b(\mathbf L_t)\mathbf Z_t
\approx
\sum_{p=0}^{P_{\mathrm{cheb}}}
 c_{b,p}T_p(\widetilde{\mathbf L}_t)\mathbf Z_t,
\  b\in\mathcal B ,
\label{eq:shared-chebyshev}
\end{equation}
where $T_p(\cdot)$ is the $p$-th Chebyshev polynomial, with $T_0(\widetilde{\mathbf L}_t)\mathbf Z_t=\mathbf Z_t$, $T_1(\widetilde{\mathbf L}_t)\mathbf Z_t=\widetilde{\mathbf L}_t\mathbf Z_t$, and the standard recurrence for $p\ge2$.
The band-specific coefficients $c_{b,p}$ differ, while the sparse polynomial bases are computed once and reused across all bands.
This preserves multi-scale filtering without per-step eigendecomposition.

Separating the spectrum is not enough, because the useful scale varies across zones and time and a fixed average would mix the signals again.
To relate each scale to the local fleet imbalance, a node-wise gate computes
\begin{equation}
\begin{aligned}
\mathbf h_{i,t}
&=\mathbf W_{\mathrm{res}}\mathbf z_{i,t}
+\sum_{b\in\mathcal B}\alpha_{i,t}^b\mathbf H_{i,t}^b ,\\
\alpha_{i,t}^b
&=
\frac{
 m_t^b\exp\!\left(
 f_g^b([\mathbf H_{i,t}^b\|\psi_{i,t}\|\mathbf e_{i,t}])
 \right)
}{
 \displaystyle\sum_{b'\in\mathcal B}
 m_t^{b'}\exp\!\left(
 f_g^{b'}([\mathbf H_{i,t}^{b'}\|\psi_{i,t}\|\mathbf e_{i,t}])
 \right)
},
\end{aligned}
\label{eq:band-gate}
\end{equation}
where $\mathbf z_{i,t}$ is row $i$ of $\mathbf Z_t$, $\mathbf e_{i,t}$ contains observed external factors, and
$\psi_{i,t}=(b_{i,t}+d_{i,t}-\bar n_{i,t})/(\bar n_{i,t}+1)$ is the observed dispatch pressure.
The weight $\alpha_{i,t}^b$ measures the dispatch relevance of scale $b$, while the residual path preserves information that should not be replaced by any single scale.
The mask $m_t^b$ applies scale dropout in training while keeping one scale active; $m_t^b=1$ for all $b$ at inference.

A dispatch representation is useful only if it can produce an executable action.
A continuous action head can assign fractional vehicles or unreachable destinations, so \name\ converts predicted shortage and move value into adjacency-masked probabilities and integer flows:
\begin{equation}
\begin{aligned}
(u_{ij,t})_{j\in\mathcal N_i^+}
&=\operatorname{Allocate}\!\left(
 \bar n_{i,t},(p_{ij,t})_{j\in\mathcal N_i^+}
\right),
\end{aligned}
\label{eq:feasible-output}
\end{equation}
where $p_{ij,t}=\operatorname{Softmax}_{j\in\mathcal N_i^+}\!\left( f_\pi(\mathbf y_{ij,t}^{\pi})\right)$,
$\mathbf y_{ij,t}^{\pi}$ is the per-vehicle move return: $\mathbf y_{ij,t}^{\pi}=[\mathbf h_{i,t}\|\mathbf h_{j,t}\| \widehat g_{j,t}-\widehat g_{i,t}\|\widehat\rho_{ij,t}]$,  $\widehat\rho_{ij,t}=f_\rho([\mathbf h_{i,t}\|\mathbf h_{j,t}\|\cdot\widehat g_{j,t}-\widehat g_{i,t}\|\tau_{ij,t}\|c_{ij,t}])$,
$\widehat g_{i,t}$ is the predicted fleet gap $\widehat g_{i,t}=b_{i,t}+d_{i,t}+\mathbf 1_H^\top\widehat{\mathbf d}_{i,t}-\mu_i\bar n_{i,t}$, and
$\widehat{\mathbf d}_{i,t}$ is the demand forecast $\widehat{\mathbf d}_{i,t}
=\operatorname{softplus}(\mathbf W_d\mathbf h_{i,t}+\mathbf b_d)$.
$\mathcal N_i^+$ contains zone $i$ and its reachable neighbors, and $\mu_i$ is the expected demand served by one available vehicle over the $H$-step horizon.
The masked softmax assigns zero probability to unreachable destinations, while $\operatorname{Allocate}$ rounds $\bar n_{i,t}p_{ij,t}$ by largest remainders.
Consequently, Equation~\eqref{eq:feasible-output} produces nonnegative, integer, adjacency-valid, and fleet-conserving actions.

Training the prediction and policy heads separately could favor scales that fit demand but do not improve rebalancing.
We therefore train the representation and dispatch policy jointly:
\begin{equation}
\begin{aligned}
\mathcal L_{\mathrm{joint}}
={}&\lambda_d\mathcal L_{\mathrm{demand}}
+\lambda_\rho\mathcal L_{\mathrm{return}}
+\lambda_\pi\mathcal L_{\mathrm{policy}}\\
&+\lambda_{\mathrm{bal}}\sum_{b\in\mathcal B}
\left(\overline\alpha^b-\frac{1}{K}\right)^2 ,
\end{aligned}
\label{eq:joint-loss}
\end{equation}
where $\overline\alpha^b$ is the gate weight averaged over training zones and times.
The demand and return losses provide direct supervision, while the clipped policy loss aligns the representation to dispatch reward.
Together with scale dropout, the weak balance term prevents the gate from collapsing before the task losses reveal which scale mixture supports rebalancing.
The learned scale features and gate weights also provide dispatch evidence for DGLS.

\subsection{Drift-Guided Layer-Selective Optimization (DGLS)}
\label{sec:dgls}

The preceding representation supports current rebalancing, yet a policy trained on old data can become unreliable when mobility patterns change.
Full-model updates may overwrite stable knowledge, while fixed parameter subsets may miss the layers affected by the current drift.
The scale features expose changes that matter to dispatch, and layer signals reveal where adaptation is needed.
Only by combining these signals with budgeted selection and drift-aware updates can the model adapt without disturbing stable knowledge.
Therefore, we propose DGLS to measure Dispatch-weighted Spectral Drift, update affected layers under a resource budget, and validate each candidate before deployment.

DGLS should not adapt to every traffic change, because a global drift score may rise even when fleet decisions are unaffected.
DGLS instead encodes a recent window $\mathcal C_t$ and a time-, zone-, and demand--supply-matched reference $\mathcal R$ with the last accepted model $\theta^\star$, while retaining each sample's graph version.
For the resulting scale sets $\mathcal C_t^b$ and $\mathcal R_t^b$, DGLS measures dispatch-weighted maximum mean discrepancy (MMD) and applies a hysteretic trigger:
\begin{equation}
\begin{aligned}
z_t
=\operatorname{Hyst}(D_t;\tau_{\mathrm{on}},\tau_{\mathrm{off}},z_{t-1}),
\end{aligned}
\label{eq:dispatch-drift}
\end{equation}
where $D_t
=\sum_{b\in\mathcal B}\overline\alpha_t^b
\widehat{\operatorname{MMD}}^2(\mathcal C_t^b,\mathcal R_t^b)$, $\overline\alpha_t^b$ is the gate weight of scale $b$ averaged over the zones and times in $\mathcal C_t$.
The operator $\operatorname{Hyst}$ activates adaptation when $D_t\ge\tau_{\mathrm{on}}$, deactivates it when $D_t\le\tau_{\mathrm{off}}$, and otherwise retains $z_{t-1}$.
The gate average removes dispatch-irrelevant changes, and $\tau_{\mathrm{on}}>\tau_{\mathrm{off}}$ avoids boundary-trigger noise.

While $D_t$ determines when adaptation is necessary, it does not specify which layers should be updated.
Updating all layers would not only exhaust the adaptation budget but could also overwrite stable knowledge.
DGLS therefore identifies the layers most relevant to the current drift by jointly considering activation drift, current gradient sensitivity, and historical importance.
Let $\mathscr A_l(\mathcal B;\theta)$ denote the activation set produced by layer $l$ for batch $\mathcal B$.
Using min--max normalization $\mathcal N_l$ across layers, the three layer signals are
$
A_l=\mathcal N_l\!\left(
\widehat{\operatorname{MMD}}^2(\mathscr A_l(\mathcal C_t^{\mathrm{tr}};\theta^\star), \mathscr A_l(\mathcal R_t;\theta^\star))\right)$,
$
G_l
=\mathcal N_l\!\left(
\frac{\|\nabla_{\theta_l}\mathcal L_{\mathrm{recent}}\|_F}
{\sqrt{|\theta_l|}}\right)$, and
$
\Omega_l
=\mathcal N_l\!\left(
\frac{1}{|\theta_l|}\sum_j[\widehat{\boldsymbol\Omega}_{l,r}]_j
\right)$.
Here, $\mathcal L_{\mathrm{recent}}$ is the joint loss on the current adaptation prefix, 
and $\widehat{\boldsymbol\Omega}_{l,r}$ is the bias-corrected moving average of squared validation gradients from accepted update $r$.

DGLS then scores the layers and selects a positive-score subset within budget $B$:
\begin{equation}
\begin{aligned}
\mathcal S_t
=\operatorname{GreedyBudget}_B
\!\left(\{(s_l,c_l)\}_{s_l>0}\right),
\end{aligned}
\label{eq:layer-selection}
\end{equation}
where $s_l
=\omega_AA_l+\omega_GG_l-\omega_I\Omega_l$, $A_l$ locates changed representations, $G_l$ measures their current effect on the objective, and $\Omega_l$ protects parameters that supported accepted behavior.
The cost $c_l$ can represent parameters, FLOPs, or execution time; $\operatorname{GreedyBudget}_B$ ranks positive-score layers by $s_l/c_l$ and adds a layer only when its cost fits the remaining budget, ensuring $\sum_{l\in\mathcal S_t}c_l\le B$.
All other parameters and optimizer states remain frozen.

Layer selection limits where adaptation occurs; preserving accepted behavior still requires protection against overfitting the short drift window.
The selected layers therefore minimize
\begin{equation}
\begin{aligned}
\mathcal L_{\mathrm{adapt}}
&=\mathcal L_{\mathrm{recent}}+\lambda_r\mathcal L_{\mathrm{ref}}
+\lambda_a\sum_{l\in\mathcal S_t}\|\mathbf a_l-\mathbf a_l^\star\|_2^2\\
&\quad+\lambda_I\sum_{l\in\mathcal S_t}
\|\widehat{\boldsymbol\Omega}_l^{1/2}\odot
(\theta_l-\theta_l^\star)\|_2^2 ,
\end{aligned}
\label{eq:adapt-loss}
\end{equation}
where $\mathcal L_{\mathrm{recent}}$ fits the current drift, while $\mathcal L_{\mathrm{ref}}$ preserves available historical demand and return behavior.
For each selected layer, $\mathbf a_l$ is its mean candidate activation, $\mathbf a_l^\star$ is the matched stable activation, and $\widehat{\boldsymbol\Omega}_l$ stores parameter-level importance whose normalized layer average gives $\Omega_l$ in Equation~\eqref{eq:layer-selection}.
The last two terms therefore protect stable activations and parameters while the recent loss learns the drift.

The protected objective controls what is retained; a single optimizer memory can nevertheless treat a short shock and a persistent change in the same way.
For $\mathbf g_{l,k}=\nabla_{\theta_l}\mathcal L_{\mathrm{adapt}}$, DGLS schedules slow writes from the shock and persistence statistics
$\omega_t=\omega_{\max}
\frac{P_t}{P_t+\Delta_t^{\mathrm{sh}}+\epsilon}$, where $P_t
=\beta_PP_{t-1}+(1-\beta_P)D_t$ and $\Delta_t^{\mathrm{sh}}
=[D_t-D_{t-1}]_+$.

The selected parameters are then updated by combining fast and slow optimizer memories:
\begin{equation}
\theta_{l,k+1}
=\theta_{l,k}
-\eta_l\operatorname{Orth}\!\left(
\frac{\widehat{\mathbf m}_{l,k}^{f}
+\omega_t\widehat{\mathbf m}_{l,k}^{s}}
{\sqrt{\widehat{\mathbf v}_{l,k}}+\epsilon}
\right),
\  l\in\mathcal S_t ,
\label{eq:fast-slow-update}
\end{equation}
where $\widehat{\mathbf m}_{l,k}^{f}$ is the bias-corrected fast momentum updated at every inner step, $\widehat{\mathbf m}_{l,k}^{s}$ is the bias-corrected slow momentum updated only when $k-k^-\ge T_t^s$, $T_t^s
=\operatorname{clip}\!\left(
\left\lfloor
T_0^s\frac{1+\kappa_S\Delta_t^{\mathrm{sh}}}{1+\kappa_PP_t}
\right\rceil,
T_{\min}^s,T_{\max}^s\right)$, and $\widehat{\mathbf v}_{l,k}$ is the bias-corrected second moment.
The index $k^-$ marks the preceding slow write, and $\eta_l$ is the layer-specific learning rate.
A sudden increase in drift enlarges $T_t^s$ and suppresses $\omega_t$, whereas persistent drift shortens the interval and increases the contribution of slow memory.
$\operatorname{Orth}$ applies short Newton--Schulz orthogonalization only to matrix directions and is the identity map for vector parameters.
Equation~\eqref{eq:fast-slow-update} therefore reacts quickly without storing a temporary spike as lasting knowledge; the complete moment recurrences are given in Appendix~\ref{app:algorithm}.

Even this protected, budgeted update remains a candidate, because limited recent data can still make it reduce dispatch reward or violate service constraints.
Candidate validation therefore uses an adaptation prefix $\mathcal C_t^{\mathrm{tr}}$ and a later held-out suffix $\mathcal C_t^{\mathrm{val}}$, both ending before decision time $t$.
The candidate and stable models are replayed from the same fleet state under identical requests and travel times, isolating the effect of the model update.
For the validation reward $R_{\mathrm{val}}$ and the lower-is-better
violation metric $q_j$, define
$
\Delta R_t
=
R_{\mathrm{val}}(\boldsymbol\theta^{\mathrm{cand}})
-
R_{\mathrm{val}}(\boldsymbol\theta^\star)$ and $
\Delta q_{j,t}
=
q_j(\boldsymbol\theta^{\mathrm{cand}})
-
q_j(\boldsymbol\theta^\star)
$.
The candidate update is accepted only when
\begin{equation}
\operatorname{Acc}_t
=
\mathbb{I}\!\left[
\Delta R_t \ge \epsilon_R
\;\land\;
\Delta q_{j,t}\le\epsilon_j,\ \forall j
\right],
\label{eq:candidate_validation}
\end{equation}
where $\epsilon_R>0$ is the required validation-reward margin and
$\epsilon_j\ge0$ is the allowed degradation tolerance for monitored
service or safety requirement $j$.

The positive reward margin prevents updates from being accepted due
to negligible or random validation fluctuations, while the constraint
tolerances reject reward-improving candidates that excessively worsen
monitored service or safety requirements.
Acceptance commits the
candidate parameters and reference statistics. 
Rejection restores the
stable model and optimizer state.

Online deployment also requires controlled computation and memory.
The shared Chebyshev basis, bounded reference buffer, layer budget, and inner-step cap bound online memory and adaptation work.
Candidate validation changes the deployed state only when Equation~\eqref{eq:candidate_validation} is satisfied.

For completeness, Appendix~\ref{app:algorithm} presents the end-to-end algorithm, auxiliary recurrences, and the online training process for the graph-wavelet and DGLS components.

\section{Experiments}
\label{sec:experiments}

We organize the evaluation around five research questions:
\begin{itemize}[left=0pt,itemsep=1pt,topsep=2pt]
    \item \textbf{RQ1}: How does \name\ compare with traditional, recent learning-based, and language-model-assisted dispatch methods?
    \item \textbf{RQ2}: How reliably does \name\ detect, adapt to, and recover from different mobility drifts?
    \item \textbf{RQ3}: What is the contribution of each proposed design to dispatch quality and continual adaptation?
    \item \textbf{RQ4}: What is DGLS's updated-parameter footprint?
    \item \textbf{RQ5}: How sensitive is \name\ to its graph-scale, layer-budget, drift-trigger, and slow-memory settings?
\end{itemize}

\subsection{Dataset}
\label{subsec:dataset}

We use two real mobility traces and one controlled simulator.
Manhattan contains 6,317 requests and 84,000 trajectory records from 350 taxis over four hours in 19 subareas, providing a compact real-city setting.
Hangzhou covers 30 days, 928 subareas, 9,041 taxis, and more than 15 million requests, and therefore tests a much larger spatial and temporal scale.
Simulate uses a $20\times20$ grid and time-varying Poisson arrivals.
Its \emph{demand rate} is the expected number of new requests per simulator step before periodic and drift multipliers are applied.
Known drift onset and recovery times are hidden from policies and used only for evaluation.
\autoref{tab:datasets} reports dataset statistics for all three settings.

\begin{table*}[t]
\centering
\normalsize
\caption{Dataset statistics.}
\Description{Statistics for the Manhattan, Hangzhou, and Simulate datasets, including duration, requests, fleet or mobility records, spatial coverage, and sampling unit.}
\label{tab:datasets}
\setlength{\tabcolsep}{4pt}
\begin{tabular*}{\textwidth}{@{\extracolsep{\fill}}lccccc@{}}
\toprule
\textbf{Dataset}
& \makecell{\textbf{Temporal}\\\textbf{span}}
& \textbf{Requests}
& \makecell{\textbf{Fleet /}\\\textbf{records}}
& \makecell{\textbf{Spatial}\\\textbf{record}}
& \makecell{\textbf{Sampling}\\\textbf{unit}} \\
\midrule
Manhattan & 4 hours & 6,317 & 350 / 84,000
& 19 subareas, $18\,\mathrm{km}^{2}$ & Second \\
Hangzhou & 30 days & 15,144,840 & 9,041 / 781,142,400
& 928 subareas, $900\,\mathrm{km}^{2}$ & Minute \\
Simulate & 800 steps & Poisson arrivals & 60 vehicles
& $20\times20$ grid & One simulator step \\
\bottomrule
\end{tabular*}
\end{table*}

\subsection{Experimental Settings}
\label{subsec:exp_settings}

\noindent\textbf{Baselines.}\quad
Traditional baselines are DGS~\cite{cheng2018dgs} and A-RTRS~\cite{riley2020artrs}.
General RL baselines include TD3+BC~\cite{fujimoto2021td3bc}, CQL~\cite{kumar2020cql}, and Decision Transformer (DT)~\cite{chen2021dt}.
Dispatch-specific recent methods are NondBREM~\cite{zhang2024nondbrem}, GARLIC~\cite{han2025garlic}, CoopRide~\cite{wang2025coopride}, and Triple-BERT~\cite{zhao2026triplebert}.
We also evaluate Q policies guided by Qwen3.5:2B or Qwen3.5:9B: the language model supplies an action prior and is not a component of \name.
Encoder comparisons replace our first module with ChebNet~\cite{defferrard2016chebnet}, GWNN~\cite{xu2019gwnn}, or WaveNet~\cite{yang2024wavenet} while retaining the same dispatch head.
For online adaptation, we compare a frozen policy, full and last-layer tuning, TENT~\cite{wang2021tent}, LoRA~\cite{hu2022lora}, PALM~\cite{maharana2025palm}, PeTTA~\cite{hoang2024petta}, and the fixed M3 optimizer (one of the key components in ~\cite{behrouz2026nested}) inspired by multi-time-scale learning~\cite{behrouz2025nested}.
The M3 updates all layers every eight dispatch steps and fixes its slow interval and weight to 8 and 0.35, without DGLS drift scoring, budget selection, importance protection, or candidate validation.

\noindent\textbf{Evaluation metrics.}\quad
The primary fleet-level metric is the empty-loaded rate, defined as $N_{\mathrm{empty}}/N_{\mathrm{total}} \times 100\%$, where $N_{\mathrm{empty}}$ denotes the number of active vehicle-time steps without passengers, including idle, pickup, and rebalancing steps, and $N_{\mathrm{total}}$ denotes all non-offline vehicle-time steps.
We additionally report passenger waiting time and operational profit, where profit is calculated as passenger revenue minus the costs of pickup, occupied travel, and rebalancing.
Adaptation performance is evaluated using drift-detection delay, the number of false triggers, recovery steps, and forgetting on a matched historical replay.
Recovery is defined as the first post-drift step at which the rolling dispatch metric returns to within 5\% of its matched no-drift value, while forgetting measures the post-update reward degradation on the historical replay.
Tables~2--4 mark column-best and second-best values in bold and underline, respectively; Table~5 marks only the best.

\noindent\textbf{Implementation details.}\quad
The real traces are divided chronologically in a 6:3:1 ratio, and Simulate uses 800,000-step streams with matched requests, fleet initialization, topology, and travel times whenever a factor is not being changed.
Sudden drift creates a short local demand or travel-time shock, whereas gradual drift moves the commuting distribution smoothly toward a shifted pattern.
Structural drift changes road connections or vehicle travel times, recurring drift removes and later restores an event pattern, and supply-side drift temporarily reduces vehicle availability.
Every drift stream has a no-drift control with the same seed and the same exogenous events outside the factor being tested.
The known onset and recovery times are stored only by the evaluator, so no adaptive method receives a drift boundary.
Candidate validation also uses only a causal held-out suffix whose outcomes are available before the current decision.
All stochastic comparisons use ten independent runs with matched random seeds, and values are reported as means, with standard deviations shown when available.
Paired bootstrap intervals and a two-sided paired permutation test at $p<0.05$ are used against the strongest baseline.
Appendix~\ref{app:experimental-details} lists all hyperparameters, drift construction, and replay rules.

\subsection{Overall Performance (RQ1)}
\label{subsec:exp_overall}

We compare \name\ with eleven traditional, learning-based, dispatch-specific, and language-model-guided policies.
\autoref{tab:garlic_dispatch} reports ten-run rates across datasets varying in size, duration, and fleet scale.
\name\ ranks first by mean with $30.22\%\pm2.04\%$, $38.54\%\pm2.33\%$, and $29.19\%\pm1.36\%$, respectively.
Relative to the strongest competing mean, CoopRide on Manhattan and Simulate, and GARLIC on Hangzhou, the reductions are 6.58\%, 5.33\%, and 3.60\% respectively.
Causal summaries and graph-wavelet bands retain broad and local patterns, while gating favors scales that improve relocation under changing regimes.

\begin{table}[t]
\centering
\normalsize
\caption{Empty-loaded rate on Manhattan (M), Hangzhou (H), and Simulate (S). Results are means${}\pm{}$standard deviations. The lower, the better.}
\Description{Ten-run empty-loaded-rate results on three datasets for traditional methods, deep-learning methods, language-model-assisted policies, and the proposed method.}
\label{tab:garlic_dispatch}
\setlength{\tabcolsep}{3.5pt}
\begin{tabular*}{\columnwidth}{@{\extracolsep{\fill}}lccc@{}}
\toprule
\textbf{Method} & \multicolumn{3}{c}{\textbf{Empty-loaded rate (\%)} $\downarrow$} \\
\cmidrule(lr){2-4}
& \textbf{M} & \textbf{H} & \textbf{S} \\
\midrule
\multicolumn{4}{@{}l}{\textbf{Traditional}} \\
DGS        & 32.57{\scriptsize$\pm$1.23} & 41.23{\scriptsize$\pm$2.85} & 30.55{\scriptsize$\pm$0.65} \\
A-RTRS     & 32.39{\scriptsize$\pm$1.37} & 41.04{\scriptsize$\pm$2.88} & 30.40{\scriptsize$\pm$1.34} \\
\midrule
\multicolumn{4}{@{}l}{\textbf{Deep Learning}} \\
TD3+BC     & 37.22{\scriptsize$\pm$3.73} & 50.13{\scriptsize$\pm$4.25} & 35.85{\scriptsize$\pm$1.96} \\
CQL        & 35.17{\scriptsize$\pm$4.66} & 46.87{\scriptsize$\pm$5.08} & 33.75{\scriptsize$\pm$2.08} \\
DT         & 33.49{\scriptsize$\pm$2.27} & 41.45{\scriptsize$\pm$2.95} & 31.05{\scriptsize$\pm$1.51} \\
NondBREM   & 33.27{\scriptsize$\pm$2.08} & 41.65{\scriptsize$\pm$3.11} & 30.85{\scriptsize$\pm$1.34} \\
GARLIC (GPT) & 32.38{\scriptsize$\pm$1.76} & \underline{40.71{\scriptsize$\pm$1.86}} & 30.31{\scriptsize$\pm$1.14} \\
CoopRide   & \underline{32.35{\scriptsize$\pm$1.28}} & 40.87{\scriptsize$\pm$1.30} & \underline{30.28{\scriptsize$\pm$1.07}} \\
Triple-BERT & 32.46{\scriptsize$\pm$1.64} & 42.44{\scriptsize$\pm$1.79} & 30.35{\scriptsize$\pm$1.22} \\
Qwen3.5-2B-guided Q & 32.60{\scriptsize$\pm$2.35} & 40.99{\scriptsize$\pm$3.03} & 30.48{\scriptsize$\pm$1.55} \\
Qwen3.5-9B-guided Q & 32.55{\scriptsize$\pm$1.91} & 40.91{\scriptsize$\pm$2.07} & 30.45{\scriptsize$\pm$1.32} \\
\midrule
\multicolumn{4}{@{}l}{\textbf{Ours}} \\
\name & \textbf{30.22{\scriptsize$\pm$2.04}} & \textbf{38.54{\scriptsize$\pm$2.33}} & \textbf{29.19{\scriptsize$\pm$1.36}} \\
\bottomrule
\end{tabular*}
\vspace{-1cm}
\end{table}

\subsection{Adaptation under Mobility Drift (RQ2)}
\label{subsec}

We compare nine adaptation methods under a no-drift setting and five matched shifts in demand, traffic, topology, recurring events, and vehicle supply, as shown in \autoref{tab:drift_robustness}.
DGLS achieves the best result in the no-drift setting and across all drift types.
Its average empty-loaded rate over the six settings is $28.53\%\pm0.64\%$, compared with $29.65\%\pm0.71\%$ for PeTTA, corresponding to a relative improvement of 3.78\%.
Compared with the best competing method, DGLS improves the empty-loaded rate by 0.99 percentage points under recurring drift and by up to 2.10 percentage points under gradual drift.
It also achieves $27.35\%\pm0.73\%$ under structural drift.
Unlike methods based on fixed update schedules, DGLS uses dispatch-related evidence to measure spectral changes, updates only the affected layers, and combines fast and slow memory with update checks.
These designs help the model adapt to mobility drift while avoiding unnecessary updates to the full model.

\begin{table*}[t]
\centering
\normalsize
\caption{Empty-loaded rate (\%) under a matched no-drift control and five mobility-drift families. Results are means${}\pm{}$standard deviations. The lower, the better.}
\Description{Ten-run empty-loaded-rate results for nine adaptation policies under a no-drift control and five mobility-drift families.}
\label{tab:drift_robustness}
\setlength{\tabcolsep}{3.5pt}
\begin{tabular*}{\textwidth}{@{\extracolsep{\fill}}lccccccc@{}}
\toprule
\textbf{Policy}
& \makecell{\textbf{No}\\\textbf{drift}}
& \textbf{Sudden}
& \textbf{Gradual}
& \textbf{Structural}
& \textbf{Recurring}
& \makecell{\textbf{Supply}\\\textbf{side}}
& \textbf{Mean} \\
\midrule
Frozen policy & 32.33{\scriptsize$\pm$1.37} & 31.30{\scriptsize$\pm$0.72} & 31.27{\scriptsize$\pm$0.92} & 30.31{\scriptsize$\pm$0.61} & 32.44{\scriptsize$\pm$1.08} & 30.85{\scriptsize$\pm$0.87} & 31.42{\scriptsize$\pm$0.84} \\
Full tuning & 31.78{\scriptsize$\pm$1.24} & 31.12{\scriptsize$\pm$0.5} & 31.03{\scriptsize$\pm$0.67} & 30.11{\scriptsize$\pm$0.62} & 31.64{\scriptsize$\pm$0.88} & 30.88{\scriptsize$\pm$0.65} & 30.79{\scriptsize$\pm$0.65} \\
Last-layer tuning & 31.28{\scriptsize$\pm$1.33} & 30.92{\scriptsize$\pm$0.87} & 31.52{\scriptsize$\pm$0.82} & 30.08{\scriptsize$\pm$0.72} & 31.31{\scriptsize$\pm$0.86} & 30.82{\scriptsize$\pm$0.72} & 30.65{\scriptsize$\pm$0.75} \\
TENT & 33.59{\scriptsize$\pm$2.03} & 31.98{\scriptsize$\pm$1.25} & 31.96{\scriptsize$\pm$0.86} & 30.55{\scriptsize$\pm$0.69} & 32.56{\scriptsize$\pm$1.17} & 31.18{\scriptsize$\pm$0.94} & 31.65{\scriptsize$\pm$0.87} \\
LoRA & 30.65{\scriptsize$\pm$1.42} & 30.98{\scriptsize$\pm$0.67} & 31.44{\scriptsize$\pm$0.85} & 30.46{\scriptsize$\pm$0.74} & 31.51{\scriptsize$\pm$0.82} & 30.79{\scriptsize$\pm$0.66} & 31.20{\scriptsize$\pm$0.64} \\
PALM & 30.45{\scriptsize$\pm$1.33} & \underline{30.67{\scriptsize$\pm$0.38}} & 31.02{\scriptsize$\pm$0.89} & 30.04{\scriptsize$\pm$0.77} & 31.24{\scriptsize$\pm$0.76} & 30.36{\scriptsize$\pm$0.62} & 30.51{\scriptsize$\pm$0.89} \\
PeTTA & \underline{30.22{\scriptsize$\pm$1.36}} & 30.72{\scriptsize$\pm$0.19} & \underline{30.55{\scriptsize$\pm$0.91}} & \underline{29.31{\scriptsize$\pm$0.82}} & \underline{30.54{\scriptsize$\pm$0.79}} & \underline{29.77{\scriptsize$\pm$0.58}} & \underline{29.65{\scriptsize$\pm$0.71}} \\
M3 only & 30.38{\scriptsize$\pm$1.32} & 30.68{\scriptsize$\pm$0.28} & 30.96{\scriptsize$\pm$1.23} & 30.04{\scriptsize$\pm$0.71} & 32.05{\scriptsize$\pm$0.93} & 30.44{\scriptsize$\pm$0.91} & 31.04{\scriptsize$\pm$0.83} \\
\textbf{DGLS (\name)} & \textbf{29.19{\scriptsize$\pm$1.36}} & \textbf{28.66{\scriptsize$\pm$0.50}} & \textbf{28.45{\scriptsize$\pm$0.79}} & \textbf{27.35{\scriptsize$\pm$0.73}} & \textbf{29.55{\scriptsize$\pm$0.69}} & \textbf{28.02{\scriptsize$\pm$0.71}} & \textbf{28.53{\scriptsize$\pm$0.64}} \\
\bottomrule
\end{tabular*}
\end{table*}

\subsection{Ablation Study (RQ3)}
\label{subsec:exp_components}

We replay single-component variants over six matched streams.
\autoref{tab:ablation} reports ten-run empty-loaded rate, profit, and waiting time to cover utilization and service.

Full \name\ obtains $29.19\%\pm1.36\%$, $79.16\pm0.53$ thousand, and $18.22\pm0.45$ steps.
Removing causal summaries raises the empty-loaded rate by 0.98 points, reduces profit by 0.67 thousand, and adds 1.07 waiting steps.
Replacing graph wavelets with a GCN raises the rate to 30.42\% and waiting to 19.73, while removing dispatch-aware gating raises them to 29.51\% and 19.51.
Causal context, scale separation, and dispatch-aware fusion are complementary.

The adaptation ablations expose metric trade-offs rather than a uniform ranking.
Unweighted drift increases profit to 80.31 thousand but also raises the empty-loaded rate to 30.11\%; removing importance protection lowers that rate to 28.92\% but increases waiting to 18.92.
Thus, aggressive short-horizon updates may improve one objective by sacrificing service or stability.
These safeguards restrain risky updates.
The full model therefore gives the best wait and balances competing objectives.

\setlength{\dblfloatsep}{6pt plus 1pt minus 1pt}
\setlength{\dbltextfloatsep}{8pt plus 1pt minus 2pt}
\begin{table*}[t]
\centering
\normalsize
\caption{Ten-run ablations over six mobility streams (mean ${}\pm{}$ standard deviation).}
\Description{Ten-run empty-loaded rate, profit, and waiting time for the full method and eight ablations.}
\label{tab:ablation}
\setlength{\tabcolsep}{4pt}
\begin{tabular*}{0.78\textwidth}{@{\extracolsep{\fill}}lccc@{}}
\toprule
\textbf{Variant}
& \makecell{\textbf{Empty-loaded} \textbf{rate (\%)} $\downarrow$}
& \makecell{\textbf{Profit} \textbf{($10^3$)} $\uparrow$}
& \makecell{\textbf{Average} \textbf{wait} $\downarrow$} \\
\midrule
\textbf{Full \name} & \underline{29.19{\scriptsize$\pm$1.36}} & 79.16{\scriptsize$\pm$0.53} & \textbf{18.22{\scriptsize$\pm$0.45}} \\
w/o causal multi-horizon summaries & 30.17{\scriptsize$\pm$0.98} & 78.49{\scriptsize$\pm$0.22} & 19.29{\scriptsize$\pm$0.84} \\
Graph wavelets $\rightarrow$ GCN & 30.42{\scriptsize$\pm$0.53} & 80.01{\scriptsize$\pm$0.43} & 19.73{\scriptsize$\pm$0.44} \\
w/o dispatch-aware gating & 29.51{\scriptsize$\pm$1.06} & \underline{80.12{\scriptsize$\pm$0.61}} & 19.51{\scriptsize$\pm$0.38} \\
Unweighted input drift & 30.11{\scriptsize$\pm$1.11} & \textbf{80.31{\scriptsize$\pm$0.58}} & 18.67{\scriptsize$\pm$0.72} \\
w/o budgeted layer selection & 29.47{\scriptsize$\pm$0.76} & 79.25{\scriptsize$\pm$0.57} & \underline{18.32{\scriptsize$\pm$0.49}} \\
w/o historical-importance protection & \textbf{28.92{\scriptsize$\pm$0.78}} & 79.21{\scriptsize$\pm$0.56} & 18.92{\scriptsize$\pm$0.65} \\
w/o drift-aware fast--slow update & 29.38{\scriptsize$\pm$1.11} & 79.46{\scriptsize$\pm$0.54} & 19.34{\scriptsize$\pm$0.70} \\
w/o candidate validation & 29.75{\scriptsize$\pm$0.94} & 79.87{\scriptsize$\pm$0.69} & 19.25{\scriptsize$\pm$0.54} \\
\bottomrule
\end{tabular*}
\end{table*}

\subsection{Parameter Updates (RQ4)}
\label{subsec:exp_efficiency}

Table~\ref{tab:efficiency} shows the number of parameters updated during adaptation under one simulation setting.
DGLS updates only 62.22K parameters, compared with 135.94K for Adam, AdamW, SGD, and M3.
This saves 73.72K parameters, or 54.23\% of baseline updates.

All four baselines update the full model, so changing the optimizer alone does not reduce the updated parameter count.
DGLS instead ranks layers using dispatch-weighted drift evidence, updates a budgeted subset, and freezes the rest.
This lowers gradient and optimizer-state costs while protecting unaffected layers.
Thus, RQ2 gains come from focused adaptation, not model compression.

\setlength{\textfloatsep}{8pt plus 1pt minus 2pt}
\begin{table}[t]
\centering
\normalsize
\caption{Updated parameters per adaptation on Simulate.}
\Description{Numbers of parameters exposed to adaptation by four full-update optimizer profiles and DGLS under the same Simulate setting.}
\label{tab:efficiency}
\setlength{\tabcolsep}{30pt}
\begin{tabular}{@{}cc@{}}
\toprule
\textbf{Method}
& \makecell{\textbf{Updated parameters}\\\textbf{(K)} $\downarrow$} \\
\midrule
\makecell[c]{Adam} & 135.94 \\
\makecell[c]{AdamW} & 135.94 \\
\makecell[c]{SGD} & 135.94 \\
\makecell[c]{M3} & 135.94 \\
\textbf{DGLS} & \textbf{62.22} \\
\bottomrule
\end{tabular}
\end{table}

\subsection{Parametric Study (RQ5)}
\label{subsec:exp_parameters}

\begin{figure*}[t]
\centering
\begin{subfigure}[t]{0.235\textwidth}
\centering
\begin{tikzpicture}
\begin{axis}[
    width=\linewidth,
    height=0.13\textheight,
    xmin=0.7,xmax=5.3,
    ymin=27.5,ymax=29.2,
    xtick={1,2,3,4,5},
    xticklabels={2,3,4,5,6},
    ylabel={Empty-loaded (\%)},
    grid=major,
    ticklabel style={font=\scriptsize},
    label style={font=\scriptsize},
    scaled y ticks=false
]
\addplot+[
    color=blue,
    mark=*,
    line width=1pt,
    error bars/.cd,
    y dir=both,
    y explicit
] coordinates {
    (1,28.3225) +- (0,0.4238)
    (2,28.3021) +- (0,0.4704)
    (3,28.2575) +- (0,0.5244)
    (4,28.4608) +- (0,0.4738)
    (5,28.3389) +- (0,0.6721)
};
\end{axis}
\end{tikzpicture}
\caption{Graph scales $K$}
\label{fig:param_scales}
\end{subfigure}
\hfill
\begin{subfigure}[t]{0.235\textwidth}
\centering
\begin{tikzpicture}
\begin{axis}[
    width=\linewidth,
    height=0.13\textheight,
    xmin=0.7,xmax=5.3,
    ymin=27.5,ymax=29.2,
    xtick={1,2,3,4,5},
    xticklabels={1,5,10,20,100},
    ylabel={Empty-loaded (\%)},
    grid=major,
    ticklabel style={font=\scriptsize},
    label style={font=\scriptsize},
    scaled y ticks=false
]
\addplot+[
    color=blue,
    mark=*,
    line width=1pt,
    error bars/.cd,
    y dir=both,
    y explicit
] coordinates {
    (1,28.3779) +- (0,0.3455)
    (2,28.3908) +- (0,0.4704)
    (3,28.2333) +- (0,0.5460)
    (4,28.2463) +- (0,0.2296)
    (5,28.2276) +- (0,0.4015)
};
\end{axis}
\end{tikzpicture}
\caption{Layer budget (\%)}
\label{fig:param_budget}
\end{subfigure}
\hfill
\begin{subfigure}[t]{0.235\textwidth}
\centering
\begin{tikzpicture}
\begin{axis}[
    width=\linewidth,
    height=0.13\textheight,
    xmin=0.7,xmax=5.3,
    ymin=27.5,ymax=29.2,
    xtick={1,2,3,4,5},
    xticklabels={0.05,0.1,0.2,0.4,0.8},
    ylabel={Empty-loaded (\%)},
    grid=major,
    ticklabel style={font=\scriptsize},
    label style={font=\scriptsize},
    scaled y ticks=false
]
\addplot+[
    color=blue,
    mark=*,
    line width=1pt
] coordinates {
    (1,28.18)
    (2,28.36)
    (3,28.52)
    (4,28.52)
    (5,28.52)
};
\end{axis}
\end{tikzpicture}
\caption{Trigger $\tau_{\mathrm{on}}$}
\label{fig:param_trigger}
\end{subfigure}
\hfill
\begin{subfigure}[t]{0.235\textwidth}
\centering
\begin{tikzpicture}
\begin{axis}[
    width=\linewidth,
    height=0.13\textheight,
    xmin=0.7,xmax=5.3,
    ymin=27.5,ymax=29.2,
    xtick={1,2,3,4,5},
    xticklabels={2,4,8,16,32},
    ylabel={Empty-loaded (\%)},
    grid=major,
    ticklabel style={font=\scriptsize},
    label style={font=\scriptsize},
    scaled y ticks=false
]
\addplot+[
    color=blue,
    mark=*,
    line width=1pt
] coordinates {
    (1,28.38)
    (2,28.18)
    (3,28.28)
    (4,28.44)
    (5,28.52)
};
\end{axis}
\end{tikzpicture}
\caption{Base slow interval $T_0^s$}
\label{fig:param_slow}
\end{subfigure}
\caption{Ten-run sensitivity of \name\ (mean ${}\pm{}$ standard deviation where available).}
\Description{Four measured sensitivity panels report the effects of graph-scale count, layer budget, trigger threshold, and base slow-memory interval on empty-loaded rate.}
\label{fig:parameter_analysis}
\end{figure*}
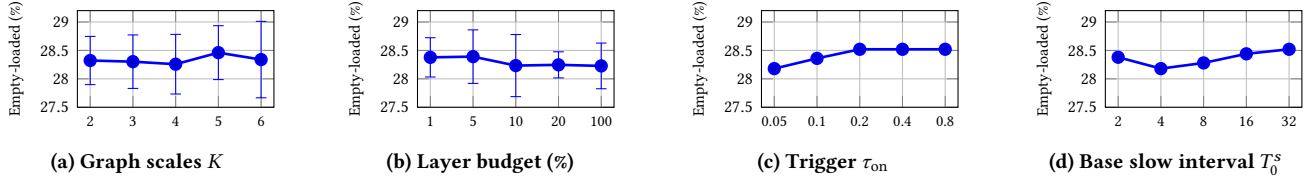

We study the effect of four key parameters by changing one parameter at a time while keeping all other training, data-stream, and dispatch settings fixed.
\autoref{fig:parameter_analysis} reports the ten-run results for the number of selected layers $K$, the layer-update budget, the drift threshold $\tau_{\mathrm{on}}$, and the slow-update interval $T_0^s$.

As shown in \autoref{fig:parameter_analysis}(a), the empty-loaded rate varies only from 28.26\% to 28.46\% when $K$ ranges from 2 to 6, with the best result at $K=4$.
A small $K$ may exclude some layers related to the current drift, while a large $K$ may update layers that still contain useful and stable knowledge.
The middle value provides enough update ability without changing too many unrelated layers.
The small overall difference also shows that the layer-ranking method can consistently identify the most useful layers.

A similar trend is observed for the layer-update budget in \autoref{fig:parameter_analysis}(b), where the results remain between 28.23\% and 28.39\%.
The 10\%, 20\%, and 100\% budgets produce similar results because drift-related changes are likely concentrated in a limited number of layers.
Once these main layers are included, increasing the budget adds little benefit and may introduce unnecessary changes to stable layers.
This explains why DGLS can achieve good performance without updating the full model.

The drift threshold has a clearer effect, as shown in \autoref{fig:parameter_analysis}(c).
Setting $\tau_{\mathrm{on}}=0.05$ achieves 28.18\%, while thresholds of 0.2 or higher give similar results of about 28.52\%.
A lower threshold allows DGLS to detect changes earlier and start adaptation before the drift causes a large loss in dispatch quality.
In contrast, a high threshold requires stronger evidence and may delay or skip useful updates.
The similar results at high thresholds suggest that these settings lead to nearly the same late-update behavior.

Finally, \autoref{fig:parameter_analysis}(d) shows that the performance first improves and then declines as $T_0^s$ increases.
The best result is 28.18\% at $T_0^s=4$, compared with 28.38\% at $T_0^s=2$ and 28.52\% at $T_0^s=32$.
When the slow memory is updated too often, short-term changes may be stored before their value is fully confirmed.
When the interval is too long, the slow memory may retain outdated information and respond too late to lasting drift.
A moderate interval therefore balances fast response and stable updates.

\section{Related Work}
\label{sec:related-work}

\subsection{Vehicle Dispatching and Rebalancing}
Vehicle dispatch has progressed from system-level guidance to learned long-horizon policies.
DGS and A-RTRS combine real-time assignment with system objectives~\cite{cheng2018dgs,riley2020artrs}, while CQL, TD3+BC, and Decision Transformer provide representative offline policy-learning strategies~\cite{kumar2020cql,fujimoto2021td3bc,chen2021dt}.
Recent methods constrain offline actions, coordinate city grids, or model driver--order relations~\cite{zhang2024nondbrem,wang2025coopride,zhao2026triplebert}.
GARLIC augments RL with multiview traffic graphs and a language-model controller~\cite{han2025garlic}.
These comparisons neither isolate dispatch-relevant graph frequencies nor select drift-responsive updates.

Human-driven systems model relocation acceptance~\cite{chen2024irebalance}.
Autonomous fleets execute feasible platform decisions directly.
This places greater weight on road representation and validated adaptation.
Frozen baseline representations cannot separate harmless input changes from harmful mobility shifts.

\subsection{Spatiotemporal Graph Learning}
Graph filters model how demand, supply, and travel conditions interact through road topology.
ChebNet evaluates localized spectral filters with sparse polynomials~\cite{defferrard2016chebnet}, and GWNN constructs graph-wavelet bases~\cite{xu2019gwnn}.
WaveNet targets nonstationary high-frequency signals~\cite{yang2024wavenet}, while WaveGC and two-dimensional filters learn richer spectral structure~\cite{liu2025wavegc,chen2025stsgnn}.
These encoder baselines optimize representation or prediction, whereas rebalancing needs a region- and time-specific measure of which graph scale changes a move decision.
\name\ learns this measure through dispatch-aware gating and reuses it to detect future mobility drift.

\subsection{Continual Adaptation}
Test-time adaptation updates a deployed model without retraining on its complete history.
TENT adapts normalization layers by entropy minimization~\cite{wang2021tent}, PALM selects layers using uncertainty and gradients~\cite{maharana2025palm}, and PeTTA limits collapse in recurring environments~\cite{hoang2024petta}.
FreqCTTA routes frequency shifts through adaptive experts~\cite{zhao2026freqctta}, while nested learning organizes optimizer memory across time scales~\cite{behrouz2025nested}.
These baselines do not tie drift evidence and update acceptance to fleet reward and service constraints.
DGLS weights spectral change by dispatch relevance, selects affected layers under a measured budget, and accepts an update only after paired candidate validation.
Thus, fleet evidence governs the trigger, layer selection, and candidate deployment under one objective.

\section{Ethical and Societal Considerations}
\label{sec:ethics}

\name\ does not introduces ethical issue.
It uses zone-level aggregate states and produces region-level fleet flows, not person-level decisions.
These routine safeguards do not affect its contributions or conclusions.

\section{Conclusion}

We presented \name\ for autonomous fleet rebalancing under evolving mobility conditions.
It combines graph wavelets for decision-relevant frequencies with DGLS, which detects drift, updates selected layers, and rejects unsafe candidate updates.
Together, the modules align representation and adaptation with fleet objectives without altering the deployed model structure.

Across ten-run evaluations, \name\ ranks first on all three datasets and DGLS ranks first under all five drift families while updating 54.23\% fewer parameters than full-model optimizers.
Sensitivity results favor a responsive drift trigger and an intermediate slow-memory interval.
Taken together, these results show that updating fewer parameters alone does not ensure reliable online adaptation.
Detected road changes must guide the choice of affected model components, and every candidate must be validated against fleet outcomes.
Future work will study directed graphs, delayed demand, battery constraints, and certified fleet deployment.


\clearpage
\bibliographystyle{ACM-Reference-Format}
\bibliography{09Reference}

\clearpage

\appendix
\setlength{\stripsep}{5pt plus 1pt minus 1pt}




\section{Complete Algorithm and Detailed Formulations}
\label{app:algorithm}

This appendix supplies implementation details that are not mentioned in the main methodology. It clarifies the causal feature boundary, sparse graph-filter evaluation, training targets, reference matching, optimizer memories, and candidate rollback. Algorithm~\ref{alg:online-adaptation} then connects these details into the online MobiWave procedure.

\subsection{Dispatch-Oriented Multi-scale Graph Wavelet Details}
\subsubsection{Causal Input Construction}
The main methodology defines the projected causal input $\mathbf Z_t$. Here we specify the features that are easy to implement inconsistently. Let $\mathbf d_t$, $\mathbf n_t$, $\bar{\mathbf n}_t=\mathbf n_t-\mathbf s_t$, and $\mathbf b_t$ stack demand, pre-assignment vehicles, remaining idle vehicles, and backlog over all zones. Define $\operatorname{cyc}_P(x)=[\sin(2\pi x/P),\cos(2\pi x/P)]$. The periodic code is $\mathbf p_t=[\operatorname{cyc}_{24}(h_t)\|\operatorname{cyc}_{7}(w_t)]$, where $h_t$ and $w_t$ are the hour-of-day and weekday indices. With $\widetilde{\mathbf W}_t=\mathbf W_t+\mathbf I$ and $[\widetilde{\mathbf D}_t]_{ii}=\sum_j[\widetilde{\mathbf W}_t]_{ij}$, the neighborhood demand--supply gap is $\boldsymbol\delta_t^{\mathrm{nbr}}=\widetilde{\mathbf D}_t^{-1}\widetilde{\mathbf W}_t(\mathbf b_t+\mathbf d_t-\bar{\mathbf n}_t)$. The arriving-vehicle feature $\mathbf a_t^{\mathrm{tr}}$ is obtained only from in-transit records already present in $\mathbf X_t$.

Let $h_{\max}=\max\mathcal H$ and $t_0=2h_{\max}-1$. The stream provides $\mathbf X_0,\ldots,\mathbf X_{t_0-1}$ as warm-up states, and the first online decision is made at $t_0$. Every temporal window used at decision time $t$ therefore ends at or before $t$, so neither the input projection nor the dispatch action uses future observations.

\subsubsection{Chebyshev Graph-Filter Evaluation}
The heat-kernel bands are evaluated without eigendecomposition. Since the normalized Laplacian has spectrum in $[0,2]$, we set $\bar\lambda=2$ and $\widetilde{\mathbf L}_t=2\mathbf L_t/\bar\lambda-\mathbf I=\mathbf L_t-\mathbf I$. For scale $b$ and Chebyshev order $P_{\mathrm{cheb}}$, the precomputed coefficient is
\begin{equation}
 c_{b,p}=\frac{2-\delta_{p0}}{\pi}\int_0^{\pi}g_b\!\left(\frac{\bar\lambda}{2}(1+\cos\vartheta)\right)\cos(p\vartheta)\,d\vartheta,\quad 0\le p\le P_{\mathrm{cheb}},
 \label{eq:app-cheb-coeff}
\end{equation}
where $\delta_{p0}$ is the Kronecker delta. The shared responses are initialized by $\mathbf R_{0,t}=\mathbf Z_t$ and $\mathbf R_{1,t}=\widetilde{\mathbf L}_t\mathbf Z_t$, and then updated by $\mathbf R_{p,t}=2\widetilde{\mathbf L}_t\mathbf R_{p-1,t}-\mathbf R_{p-2,t}$ for $p\ge2$. Every spectral scale reuses these responses with its own coefficients $c_{b,p}$. The sparse recurrence and the $K$ scale-specific weighted sums cost $O(P_{\mathrm{cheb}}(|\mathcal E_t|+KN)F_z)$. A topology change therefore rebuilds only $\widetilde{\mathbf L}_t$; the heat-scale coefficients remain fixed.

\subsubsection{Prediction and Feasible Dispatch}
In the Methodology section, we define the dispatch-aware gate and deterministic allocation rule. For reproducibility, each scale logit is produced by a one-hidden-layer ReLU network $f_g^b$ from $[\mathbf H^b_{i,t}\|\psi_{i,t}\|\mathbf e_{i,t}]$. The demand head applies softplus to a linear projection of $\mathbf h_{i,t}$, and the predicted fleet gap adds current backlog, current demand, and forecast demand before subtracting the service capacity of idle vehicles. The return head receives the endpoint representations, gap difference, travel time, and move cost; the policy head additionally receives the predicted move return.

For training, the adjacency-masked probabilities define a multinomial policy over valid destinations $\mathcal N_i^+=\{i\}\cup\{j:(v_i,v_j)\in\mathcal E_t\}$:
\begin{equation}
 \pi_\theta(\mathbf U_t\mid\mathbf X_{\le t},\mathcal G_t)=\prod_{i=1}^{N}\frac{\bar n_{i,t}!}{\prod_{j\in\mathcal N_i^+}u_{ij,t}!}\prod_{j\in\mathcal N_i^+}p_{ij,t}^{u_{ij,t}}.
 \label{eq:app-multinomial-policy}
\end{equation}

The masked softmax sets $p_{ij,t}=0$ for $j\notin\mathcal N_i^+$. During deployment, no stochastic sample is used: the model applies the deterministic largest-remainder allocation stated in the main methodology. Thus, both training and deployment preserve nonnegative integer flows, adjacency validity, and $\sum_j u_{ij,t}=\bar n_{i,t}$.

\subsubsection{Training Targets}
The main methodology gives the weighted joint objective. This subsection specifies only its supervision targets. Let $\mathcal V_{ij,t}$ be the nonempty set of vehicles moved from $i$ to $j$, $M_{j,t}=\sum_i|\mathcal V_{ij,t}|$, $R^v_{t,H}=\sum_{q=0}^{H-1}\gamma^q r^v_{t+q}$, and $C^H_{j,t}=\sum_{q=1}^{H}\gamma^{q-1}\lambda_c C_{j,t+q}$. The realized per-vehicle move return is
\begin{equation}
 \rho_{ij,t}=\frac{1}{|\mathcal V_{ij,t}|}\sum_{v\in\mathcal V_{ij,t}}R^v_{t,H}-\frac{C^H_{j,t}}{\max(1,M_{j,t})}.
 \label{eq:app-move-return-target}
\end{equation}

A target enters $\mathcal I_\rho$ only after all $H$ outcomes and the final cancellation boundary have been observed; incomplete suffixes and empty move sets remain unlabeled.

For a causal rollout batch $\mathcal Q$, the demand loss is
\begin{equation}
 \mathcal L_{\mathrm{demand}}=\frac{1}{|\mathcal Q|NH}\sum_{t\in\mathcal Q}\sum_{i=1}^{N}\sum_{r=1}^{H}\operatorname{Huber}_{\delta_d}(\widehat d_{i,t+r}-d_{i,t+r}).
 \label{eq:app-demand-loss}
\end{equation}

When $\mathcal I_\rho\neq\varnothing$, the return loss is
\begin{equation}
 \mathcal L_{\mathrm{return}}=\frac{1}{|\mathcal I_\rho|}\sum_{(t,i,j)\in\mathcal I_\rho}\operatorname{Huber}_{\delta_\rho}(\widehat\rho_{ij,t}-\rho_{ij,t}),
 \label{eq:app-return-loss}
\end{equation}
and it is set to zero otherwise. Let $G_t$ be the discounted reward-to-go within the sampled rollout, $\widehat A_t$ its batch-standardized value, and $\chi_t(\theta)$ the ratio between the current and stored behavior-policy probabilities. The clipped policy loss is
\begin{equation}
 \mathcal L_{\mathrm{policy}}=-\frac{1}{|\mathcal Q|}\sum_{t\in\mathcal Q}\min\!\left(\chi_t\widehat A_t,\operatorname{clip}(\chi_t,1-\epsilon_\pi,1+\epsilon_\pi)\widehat A_t\right).
 \label{eq:app-policy-loss}
\end{equation}

Each rollout tuple stores $\mathbf U_t$ and its behavior log probability, so the denominator of $\chi_t$ is fixed during optimization. The average gate weight used by the balance regularizer is computed over all zones and time steps in $\mathcal Q$.

\subsection{DGLS Details}

\subsubsection{Drift, Reference Matching, and Layer Statistics}
For each spectral scale, DGLS deterministically selects at most 256 feature rows from the recent and matched reference sets.
Let $\overline k(\mathcal A,\mathcal B)$ denote the average kernel value over all cross-set pairs.
We use the Gaussian kernel $k_\upsilon(\mathbf x,\mathbf y)=\exp(-\|\mathbf x-\mathbf y\|_2^2/(2\sigma_\upsilon^2))$.
The empirical discrepancy used by the drift score and layer diagnostics is
\begin{equation}
 \widehat{\operatorname{MMD}}^2(\mathcal A,\mathcal B)=\overline k(\mathcal A,\mathcal A)+\overline k(\mathcal B,\mathcal B)-2\overline k(\mathcal A,\mathcal B).
 \label{eq:app-mmd}
\end{equation}

Before deployment, $\sigma_\upsilon$ is fixed to the median nonzero pairwise distance of at most 512 deterministically selected reference rows, lower-bounded by $\epsilon_\sigma$. The kernel sums are evaluated exactly on the bounded sets in memory-bounded blocks.

The reference buffer is initialized from training history. Each recent sample is matched by time stratum, zone, and demand--supply level. If the exact stratum is empty, the closest nonempty stratum is selected lexicographically by zone-graph distance, cyclic time distance, and standardized gap difference, with disconnected zones ordered last. Recent and reference samples are encoded by the same accepted parameters $\theta^\star$ while retaining their own graph versions, which allows structural drift to be measured rather than matched away.

For candidate adaptation, $\mathcal L_{\mathrm{recent}}$ is the joint loss on $\mathcal C_t^{\mathrm{tr}}$. The reference loss contains only available demand and return labels; no policy term is computed because the compact buffer does not retain old trajectory advantages or behavior probabilities. Unlabeled reference inputs still define the stable activation anchor $\mathbf a_l^\star$, while the candidate prefix defines $\mathbf a_l$.

The main methodology defines activation drift, current gradient sensitivity, historical importance, and the budgeted layer score. The parameter-wise importance state advances only after accepted update $r$:
\begin{equation}
 \boldsymbol\Omega_{l,r}=\beta_I\boldsymbol\Omega_{l,r-1}+(1-\beta_I)(\mathbf g^{\mathrm{hist}}_{l,r})^{\odot2}.
 \label{eq:app-importance}
\end{equation}

For $r>0$, its bias-corrected value is $\widehat{\boldsymbol\Omega}_{l,r}=\boldsymbol\Omega_{l,r}/(1-\beta_I^r)$. At first deployment, $r=0$ and the bias-corrected importance is defined as zero. $\widetilde s_l=\max(0,s_l)/\{\max_q[\max(0,s_q)]+\epsilon\}$ is the normalized positive score. After layers are selected under budget $B$, their learning rates are
\begin{equation}
 \eta_l=\eta\widetilde s_l\left[\eta_{\min}+(1-\eta_{\min})(1-\Omega_l)\right],
 \label{eq:app-layer-lr}
\end{equation}
where floor $\eta_{\min}$ allows an important selected layer to move cautiously instead of forcing its learning rate to zero. All hyperparameters and the positive resource budget are fixed on the adaptation split before test-time deployment.

\subsubsection{Fast--Slow Update}
The main methodology defines the shock statistic $\Delta_t^{\mathrm{sh}}$, persistence statistic $P_t$, slow-write interval $T_t^s$, and protected candidate objective. After the diagnostic pass, only the selected layers are trainable. Before each optimizer step, their joint gradient is clipped to global $\ell_2$ norm 5.

Fast momentum and the second moment follow the usual exponential recurrences and are bias-corrected by the number of fast updates since the layer was activated. Let $k^-$ denote the preceding slow-write step and $\overline{\mathbf g}_{l,k}=(k-k^-)^{-1}\sum_{q=k^-+1}^{k}\mathbf g_{l,q}$. The retained slow memory is updated only when the scheduled interval is reached:
\begin{equation}
 \mathbf m^s_{l,k}=\begin{cases}\beta_s\mathbf m^s_{l,k^-}+(1-\beta_s)\overline{\mathbf g}_{l,k},&k-k^-\ge T_t^s,\\ \mathbf m^s_{l,k^-},&k-k^-<T_t^s.\end{cases}
 \label{eq:app-slow-write}
\end{equation}

The slow-write counter increments only in the first case, after which $k^-$ is set to $k$; otherwise the slow state and counter remain unchanged. Newly reactivated layers reset their fast momentum, second moment, and fast counter, but retain the accepted slow memory and its write counter. With bias-corrected memories, the normalized candidate direction is
\begin{equation}
 \mathbf u_{l,k}=\frac{\widehat{\mathbf m}^{f}_{l,k}+\omega_t\widehat{\mathbf m}^{s}_{l,k}}{\sqrt{\widehat{\mathbf v}_{l,k}}+\epsilon},
 \label{eq:app-fast-slow-direction}
\end{equation}
where $\omega_t=\omega_{\max}\frac{P_t}{P_t+\Delta_t^{\mathrm{sh}}+\epsilon}$. 

When the parameters are vectors, we use $\theta_{l,k+1}=\theta_{l,k}-\eta_l\mathbf u_{l,k}$. For a matrix parameter, reshape the direction to $\widetilde{\mathbf u}_{l,k}$ and transpose it when the matrix is tall. Starting from $\mathbf Y_0=\widetilde{\mathbf u}_{l,k}/(\|\widetilde{\mathbf u}_{l,k}\|_F+\epsilon)$, the short Newton--Schulz iteration is
\begin{equation}
\theta_{l,k+1}=\theta_{l,k}-\eta_l\operatorname{restore}(\mathbf Y_{Q_{\mathrm{NS}}}).
 \label{eq:newton-schulz}
\end{equation}
where $\mathbf Y_{q+1}=\tfrac32\mathbf Y_q-\tfrac12(\mathbf Y_q\mathbf Y_q^\top)\mathbf Y_q$.

Thus, a sudden change is handled mainly by fast memory, whereas persistent drift gradually contributes to the retained slow direction.

\subsubsection{Candidate Validation and Rollback}
Candidate validation uses the held-out suffix $\mathcal C_t^{\mathrm{val}}$, which follows the adaptation prefix but ends before the current decision time. The candidate and accepted policies are replayed from the same fleet state under identical requests, travel times, and graph evolution. If paired replay is unavailable, the candidate is rejected. Otherwise, the main-text acceptance criterion requires the validation reward to improve by at least $\epsilon_R$ and every lower-is-better service or safety metric to worsen by no more than its tolerance $\epsilon_j$.

Acceptance atomically commits the candidate parameters and optimizer state, advances the accepted-update index, updates parameter importance with the accepted validation gradient, and appends compact validation inputs, available targets, and graph versions to the fixed-capacity reference buffer. Rejection discards the isolated candidate and leaves the accepted parameters, optimizer memories, importance state, and reference buffer unchanged.

\subsection{Online Training Process}

The detailed online training process is given in Algorithm~\ref{alg:online-adaptation}. Its inputs are the mobility stream, the accepted model and optimizer state, the importance state and reference buffer, the hysteresis thresholds, the layer budget, and the maximum number of candidate steps, and its outputs are feasible fleet flows and the final accepted state. Line~1 initializes the drift score, persistence memory, and hysteresis state to zero before online dispatch begins. Line~2 iterates over every dispatch time from $t_0$ to $T$. Line~3 divides observations strictly earlier than $t$ into a causal adaptation prefix $\mathcal C_t^{\mathrm{tr}}$ and a later held-out validation suffix $\mathcal C_t^{\mathrm{val}}$. Line~4 uses the currently accepted parameters $\theta^\star$ and the state observed through time $t$ to produce a feasible fleet flow $\mathbf U_t$. Line~5 measures dispatch-weighted spectral drift against the accepted reference buffer and updates the binary trigger through hysteresis. Line~6 converts the current and preceding drift scores into the short-term shock statistic $\Delta_t^{\mathrm{sh}}$ and persistence statistic $P_t$. Line~7 permits adaptation only when the trigger is active and both causal windows contain enough completed observations. Line~8 ranks the affected layers, selects a subset $\mathcal S_t$ within budget $B$, and assigns their protected learning rates $\boldsymbol\eta_t$. Line~9 proceeds only when at least one layer has been selected. Line~10 copies the accepted parameters and optimizer memories into an isolated candidate state, with only the selected layers enabled for updating. Line~11 limits candidate fitting to at most $M_{\max}$ inner optimization steps. Line~12 computes the gradient of the protected adaptation objective from the causal prefix and accepted reference samples. Line~13 applies the drift-aware fast--slow update using the gradient, shock, persistence, and layer-wise learning rates, thereby changing only the selected candidate layers and their candidate optimizer state. Line~14 compares the candidate and accepted models on the same held-out suffix under matched replay and returns the acceptance indicator $\mathsf{Acc}_t$. Line~15 enters the commit branch only when the candidate satisfies the reward and service constraints. Line~16 atomically accepts the candidate parameters and optimizer state, updates historical importance, and refreshes the bounded reference buffer. Line~17 recomputes $\mathbf U_t$ with the newly accepted model so that the current action reflects an accepted update rather than an unvalidated candidate. Line~18 executes the resulting feasible flow; if adaptation was not ready, no layer fit the budget, or validation failed, this is the flow already produced by the unchanged accepted model in Line~4.

\begin{algorithm}[t]
\LinesNumbered
\small
\caption{MobiWave Online Training and Adaptation}
\label{alg:online-adaptation}
\KwIn{Stream $\{(\mathcal G_t,\mathbf X_t)\}_{t=t_0}^{T}$; accepted state $(\theta^\star,\mathcal O^\star,\boldsymbol\Omega,\mathcal R)$; thresholds $\tau_{\mathrm{on}}>\tau_{\mathrm{off}}$; budget $B$; step cap $M_{\max}$}
\KwOut{Feasible flows $\{\mathbf U_t\}$ and the final accepted state}
$(D_{t_0-1},P_{t_0-1},z_{t_0-1})\leftarrow(0,0,0)$\;
\For{$t\leftarrow t_0$ \KwTo $T$}{
 $(\mathcal C_t^{\mathrm{tr}},\mathcal C_t^{\mathrm{val}})\leftarrow\operatorname{CausalSplit}(\mathcal G_{<t},\mathbf X_{<t})$\;
 $\mathbf U_t\leftarrow\operatorname{Dispatch}_{\theta^\star}(\mathcal G_t,\mathbf X_{\le t})$\;
 $D_t\leftarrow\operatorname{SpectralDrift}_{\theta^\star}(\mathcal C_t^{\mathrm{tr}}\cup\mathcal C_t^{\mathrm{val}},\mathcal R)$; $z_t\leftarrow\operatorname{Hyst}(D_t,z_{t-1})$\;
 $(\Delta_t^{\mathrm{sh}},P_t)\leftarrow\operatorname{DriftMemory}(D_t,D_{t-1},P_{t-1})$\;
 \If{$z_t=1$ \textbf{and} $\operatorname{Ready}(\mathcal C_t^{\mathrm{tr}},\mathcal C_t^{\mathrm{val}})$}{
  $(\mathcal S_t,\boldsymbol\eta_t)\leftarrow\operatorname{SelectLayers}(\theta^\star,\mathcal C_t^{\mathrm{tr}},\mathcal R,\boldsymbol\Omega,B)$\;
  \If{$\mathcal S_t\neq\varnothing$}{
   $(\theta^{\mathrm{cand}},\mathcal O^{\mathrm{cand}})\leftarrow\operatorname{InitCandidate}(\theta^\star,\mathcal O^\star,\mathcal S_t)$\;
   \For{$k\leftarrow1$ \KwTo $M_{\max}$}{
    $\mathbf g_k\leftarrow\nabla_{\theta^{\mathrm{cand}}_{\mathcal S_t}}\mathcal L_{\mathrm{adapt}}(\mathcal C_t^{\mathrm{tr}},\mathcal R)$\;
    $(\theta^{\mathrm{cand}}_{\mathcal S_t},\mathcal O^{\mathrm{cand}})\leftarrow\operatorname{FastSlowStep}(\mathbf g_k,\Delta_t^{\mathrm{sh}},P_t,\boldsymbol\eta_t,\mathcal O^{\mathrm{cand}})$\;
   }
   $\mathsf{Acc}_t\leftarrow\operatorname{CandidateValidate}(\theta^{\mathrm{cand}},\theta^\star,\mathcal C_t^{\mathrm{val}})$\;
   \If{$\mathsf{Acc}_t=1$}{
    $(\theta^\star,\mathcal O^\star,\boldsymbol\Omega,\mathcal R)\leftarrow\operatorname{Commit}(\theta^{\mathrm{cand}},\mathcal O^{\mathrm{cand}},\mathcal C_t^{\mathrm{val}})$\;
    $\mathbf U_t\leftarrow\operatorname{Dispatch}_{\theta^\star}(\mathcal G_t,\mathbf X_{\le t})$\;
   }
  }
 }
 Execute $\mathbf U_t$\;
}
\end{algorithm}

\section{Experimental Details}
\label{app:experimental-details}

\subsection{Datasets and Preprocessing}
Manhattan contains 350 taxis, 84,000 trajectories, and 6,317 requests over four hours.
Hangzhou contains 9,041 taxis, 781,142,400 trajectories, and 15,144,840 requests over 30 days.
Both real-world datasets are divided chronologically in a 6:3:1 ratio.
Simulate is a $20\times20$ grid whose demand rate 2.4 means that 2.4 new requests are expected over the whole grid in each dispatch step before time-varying spatial and temporal changes are applied.
\autoref{tab:datasets} in the main paper gives the full statistics.

\subsection{Protocols and Implementation}
All simulator experiments use the same default protocol: a $20\times20$ grid, 60 vehicles, an 800-step horizon, demand rate 2.4, maximum wait 15, state-hop radius 3, and ten fixed random seeds.
Vehicle movement costs 0.1 per grid step, and loaded passenger travel contributes 5 per grid step to revenue.
The predictor sweep covers historical, MLP, and Graph Wavelet representations with AdamW, SGD, and the recorded DGLS configuration.
The workbook labels this configuration \texttt{m3}.
Table~\ref{tab:ablation} isolates the graph-wavelet and DGLS factors with the matched MLP and AdamW replacements.
Legacy runs with a different simulator protocol are excluded; a row enters the reported tables only when its manifest matches the stated configuration, seeds, and source hash.

The fast-momentum, slow-momentum, and second-moment factors are $(0.9,0.99,0.999)$.
The slow write interval is eight steps, the maximum slow-memory weight is 0.35, and matrix updates use four Newton--Schulz iterations.
The MobiWave backbone uses 20 offline pretraining epochs.
All compared methods receive the same request stream, fleet initialization, action constraints, and seed.
The test stream inherits the bidirectional OD counts collected in the offline history, and graph weights use feasible links and free-flow travel times; observed incident delays remain in the causal state used by the output heads.
Greedy nearest and Demand balance require no training, whereas PPO and the two Qwen policies use 30 epochs.
Run directories record the model identifiers \texttt{qwen3.5:9b} and \texttt{qwen3.5:2b}.
The operating-condition suite changes one factor at a time: fleet size in $\{30,120\}$, demand rate in $\{1.5,3.5\}$, maximum wait in $\{5,25\}$, and horizon in $\{600,1000\}$.
Qwen3.5:2B uses 20 epochs in this suite.

\subsection{Drift Construction}
\textbf{Sudden drift} introduces a localized demand or travel-time shock that represents an accident, heavy rain, or a large event.
\textbf{Gradual drift} interpolates between historical and shifted commuting patterns over an extended interval.
\textbf{Structural drift} closes or adds road connections or persistently changes selected edge travel times.
\textbf{Recurring drift} removes an event pattern and later restores it.
\textbf{Supply-side drift} changes fleet size, vehicle availability, or charging-induced downtime.
Each family uses the same replay interface and has a matched no-drift control.
Ground-truth onset and recovery times are retained only for evaluation.

\end{document}